\documentclass{article}

\usepackage{arxiv}

\usepackage[utf8]{inputenc} 
\usepackage[T1]{fontenc}    
\usepackage{hyperref}       
\usepackage{url}            
\usepackage{booktabs}       
\usepackage{amsfonts}       
\usepackage{nicefrac}       
\usepackage{microtype}      
\usepackage{lipsum}		
\usepackage{graphicx}
\usepackage{natbib}
\usepackage{doi}
\usepackage{tabularx}
\usepackage{multirow}

\newcommand\blfootnote[1]{%
  \begingroup
  \renewcommand\thefootnote{}\footnote{#1}%
  \addtocounter{footnote}{-1}%
  \endgroup
}

\title{Parkinson's disease diagnostics using AI and natural language knowledge transfer}

\author{ \href{https://scholar.google.com/citations?user=cZ2rFq8AAAAJ}{\hspace{1mm}Maurycy Chronowski} \\
	Department of Applied Computer Science \\
	AGH University of Science And Technology \\
	Mickiewicza 30, 30-059, Kraków, Poland \\
	\texttt{moryc.chronowski@gmail.com} \\
	\And
	\href{https://orcid.org/0000-0002-2009-1216}{\includegraphics[scale=0.06]{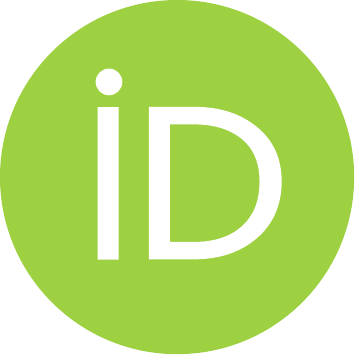}\hspace{1mm}Maciej Kłaczyński} \\
	Department of Mechanics and Vibroacoustics \\
	AGH University of Science And Technology \\
	Mickiewicza 30, 30-059, Kraków, Poland \\
	\texttt{maciej.klaczynski@agh.edu.pl} \\
	\And
	\href{https://orcid.org/0000-0002-9829-3301}{\includegraphics[scale=0.06]{orcid.pdf}\hspace{1mm}Małgorzata Dec-Ćwiek} \\
	Department of Neurology \\
	Jagiellonian University, Collegium Medicum \\
	Jakubowskiego 2, 30-688, Kraków, Poland \\
	\texttt{malgorzata.dec-cwiek@uj.edu.pl} \\
    \And
	\href{https://orcid.org/0000-0003-2494-4703}{\includegraphics[scale=0.06]{orcid.pdf}\hspace{1mm}Karolina Porębska} \\
	Department of Neurology \\
	Jagiellonian University, Collegium Medicum \\
	Jakubowskiego 2, 30-688, Kraków, Poland \\
	\texttt{karolina.porebska@uj.edu.pl} \\
}

\renewcommand{\shorttitle}{\textit{arXiv} Template}

\hypersetup{
pdftitle={Parkinson's disease diagnostics using AI and natural language knowledge transfer},
pdfsubject={q-bio.NC, q-bio.QM},
pdfauthor={Maurycy Chronowski},
pdfkeywords={Parkinson's disease, Digital diagnostics, Artificial intelligence, Speech processing},
}

\begin{document}
\maketitle

\blfootnote{Our code is publicly available at: \textit{code will be published upon paper acceptance}}


\begin{abstract}
In this work, the issue of Parkinson's disease (PD) diagnostics using non-invasive antemortem techniques was tackled. A deep learning approach for classification of raw speech recordings in patients with diagnosed PD was proposed. The core of proposed method is an audio classifier using knowledge transfer from a pretrained natural language model, namely \textit{wav2vec 2.0}. Method was tested on a group of 38 PD patients and 10 healthy persons above the age of 50. A dataset of speech recordings acquired using a smartphone recorder was constructed and the recordings were label as PD/non-PD with severity of the disease additionally rated using Hoehn-Yahr scale. The audio recordings were cut into 2141 samples that include sentences, syllables, vowels and sustained phonation. The classifier scores up to 97.92\% of cross-validated accuracy. Additionally, paper presents results of a human-level performance assessment questionnaire, which was consulted with the neurology professionals.
\end{abstract}




\section{Introduction}
Parkinson's disease (PD) is a progressive disorder of the nervous system that affects parts of the brain responsible for the motor functions. It is estimated that in industrialised societies PD affects about 1\% of the population above the age of 60 \citep{sveinbjornsdottir_clinical_2016}. Despite its commonness, there is still no antemortem test for PD. Therefore, the diagnosis relies on patient's history and physical examination. Novel approaches are examined in works such as \citep{signaevsky_antemortem_2022}. 

Previous findings, including \citep{ali_early_2019, almeida_detecting_2019, sakar_collection_2013}, have shown that Parkinson's disease can be accurately diagnosed using speech recordings and machine learning techniques. In authors' earlier research covering the presented PD dataset \citep{chronowski_speech_2020} the results indicated a significant signal in speech recordings acquired using a smartphone. In this work, we approach this topic using deep learning audio models. We propose an architecture based on wav2vec 2.0 \citep{schneider_wav2vec_2019} and we test it in a transfer learning setup. Ultimately, we discuss the possibility of implementing our approach as a remote diagnostics tool and we present a human-level performance assessment consulted with the medical experts in the neurology domain.

Our goal was to determine if an audio model that was trained on a large-scale natural language dataset can be transferred and fine-tuned to a downstream task of medical diagnostics. Medical tasks usually suffer from insufficient amount of labelled training data, therefore it would be much beneficial to observe such knowledge transfer.

\section{Related works}

\subsection{Wav2vec2.0}
As a backbone architecture for our experiments, we use pretrained parts of the wav2vec 2.0 model. It is a raw audio speech recognition transformer model published in \citep{schneider_wav2vec_2019}. The model is pretrained in an unsupervised manner and was shown to deliver state-of-the-art performance in speech recognition tasks using very limited fine-tuning. In this work, we utilise the pretrained convolutional layers of wav2vec 2.0.

\subsection{Explainability in AI}
Explainable AI (XAI) plays an important role in medical applications of artificial intelligence. It was shown in \citep{ribeiro_why_2016} that black-box models with wrong explanations encourage distrust in deep learning models, despite their good overall performance. It is therefore important to design models in a way that their predictions can be explained and understood by domain experts, who might not be familiar with machine learning at all. In this work, we discuss possible explanations of the audio models and we present a survey among neurology experts, aiming to assess the human-level performance of speech-based PD diagnostics. 

\section{Proposed method}

\subsection{Data acquisition}
The data was acquired according to the previous research presented in \citep{chronowski_speech_2020}. The dataset consisted of phonetic test recordings gathered using a mid-range Android smartphone. PD patients were labelled with Hoehn-Yahr ratings by the neurologists at the clinic and the clinical hospital. Healthy persons were recruited from participants above the age of 50, as majority of the PD patients are among the elderly. This helped to mitigate the potential age-related bias.

The patients were asked to read out loud a set of vowels (including sustained phonation), syllables, and sentences in Polish language. Full recordings were segmented into comprehensible audio fragments (a detailed list is included in Appendix \ref{app:phonetic_test}).

\subsection{Preprocessing and data augmentation}
Before entering the pipeline, two-channel smartphone recordings were subtracted from each other for noise cancellation, as described in previous work \citep{chronowski_speech_2020}. The recordings were also peak-normalised to common gain.

Taking into account the relatively small number of available audio samples in the dataset (2141) and a need for broad domain generalisation stemming from the usage of a smartphone recorder, it was necessary to strongly augment the dataset. So-called "audiomentations" \citep{iver_jordal_2022_6381721} were used, including: \textit{addition of random background noise, addition of random coloured noise, random shift in time domain, random polarity inversion}. A detailed description of the augmentations and their setup is included in Appendix \ref{app:audiomentations}. The samples were randomly augmented during each iteration of the training and were turned off during testing.

\begin{figure}[h]
	\centering
	\includegraphics[width=0.7\textwidth]{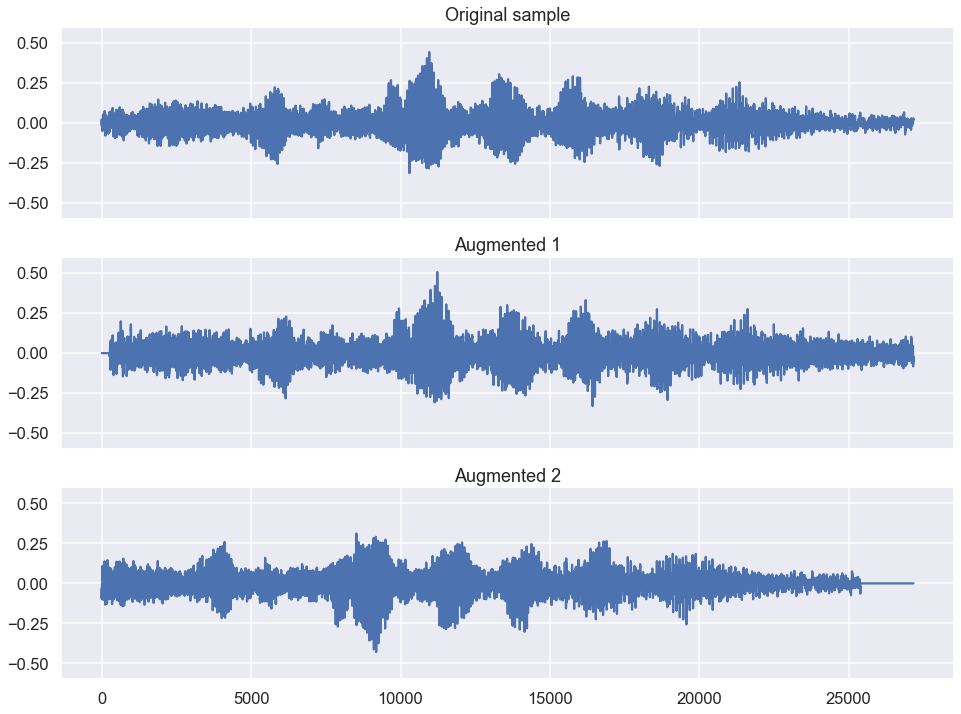}
	\caption{Visualisation of the signal waveform before (first row) and after augmentations (bottom two). Both of the augmented signals are still clearly intelligible to the human ear.}
	\label{fig:augmentations}
\end{figure}

\subsection{Model architecture}

\begin{figure}[h]
	\centering
	\includegraphics[width=\textwidth]{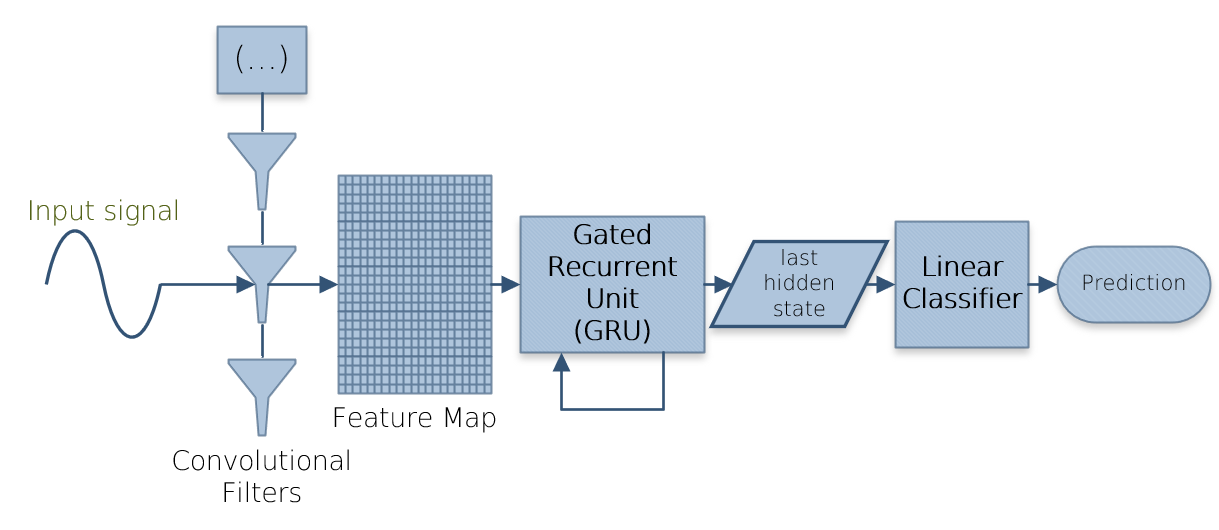}
	\caption{Proposed model architecture. The convolutional layer was taken from a pretrained wav2vec 2.0 model.}
	\label{fig:model}
\end{figure}

The model architecture was designed in a sequence-to-one manner. The input to the model was expected to be a single-channel raw audio waveform that was then internally processed into a vector representation and classified into a class label. Wav2vec 2.0 model is by design a sequence-to-sequence transformer, therefore, sequence aggregation had to be performed after the representation was obtained.

Among tested configurations, the best-performing one was a GRU that was using a convolutional feature map from wav2vec as the input. We tested also a full transformer setup, but it failed to converge in every experimental run. Training logs from both of described approaches are shown in Appendix \ref{app:training}. The last hidden state of the GRU layer was passed on to a linear classifier that generated per-sample predictions.

\subsection{Voting inference}
The models were trained to classify segmented audio samples. However, the final prediction needs
to aggregate all of the single-sample predictions for a given patient. Using an end-to-end Multiple Instance Learning setup \citep{babenko_multiple_2008} was restricted due to hardware limitations. Voting inference was proposed to counteract this obstacle.

After the models were trained to classify single samples, their predictions were aggregated for each patient. The final output label was the mode value of single-sample predictions. In the results, we report both the single-sample and aggregated voting performance.

\subsection{Experimental setup}
\label{subsec:setup}
In our dataset, we gathered 38 PD patients at different stages of the disease's development (a detailed Hoehn-Yahr table is presented in Appendix \ref{app:HY}) and 10 healthy persons (HP) above the age of 50. After segmentation, the dataset consisted of a total of 2141 audio samples ranging from vowels to full sentences. Audio had to be resampled from original 44.1 kHz sampling rate to 16 kHz, which is the sampling frequency using which the wav2vec backbone was trained \citep{schneider_wav2vec_2019}. 

To verify the hypothesis that knowledge from pretrained natural language audio models can be transferred to medical tasks, we trained our models in 3 configurations:
\begin{itemize}
    \item baseline model with pretrained and frozen convolutional layers (frozen conv)
    \item baseline model with pretrained convolutional layers and full fine-tuning (full + pretrained)
    \item baseline model with randomly initialised layers and full training (full + not pretrained)
\end{itemize}

The pretrained model that we used was Wav2Vec 2.0 Base with no fine-tuning. The GRU was a bidirectional unit with 1 hidden layer and hidden size 256. Classifier head consisted of 2 hidden layers with hidden size equal to 128.

Each of the configurations was trained in a 5-fold cross-validation setup. The folds were stratified in terms of Hoehn-Yahr score, meaning that each fold contained patients at different stages of PD. The reported metrics were averaged across the folds.

Models were trained for 400 epochs with batch size 32 and Adam optimizer with 10e-4 learning rate and betas equal to (0.9, 0.999) on a Nvidia Tesla K40 XL GPU.

\section{Results}
The results below are presented for a setup described in \ref{subsec:setup}, unless otherwise noted. We report averaged 5-fold cross-validated test metrics.

\begin{center}
\begin{table}[h!]
\centering
\caption{Comparison of models' scores between different training schemes}
\setlength{\arrayrulewidth}{1pt}
\begin{tabularx}{1\textwidth} { 
  | >{\centering\arraybackslash}X 
  | >{\centering\arraybackslash}X 
  | >{\centering\arraybackslash}X
  | >{\centering\arraybackslash}X | }
 \hline
 Model & Single-sample accuracy & Inferred voting accuracy & Inferred voting ROC AUC \\ \hline \hline
 frozen conv  & \textbf{89.92\%} & \textbf{97.92\%} & \textbf{0.99} \\ \hline
 full + pretrained & 82.07\% & 83.33\% & 0.84 \\ \hline
 full + not pretrained & 81.11\% & 83.33\% & 0.80 \\ \hline
\end{tabularx}
\label{tab:results}
\end{table}
\end{center}

\begin{figure}[h]
	\centering
	\includegraphics[width=0.6\textwidth]{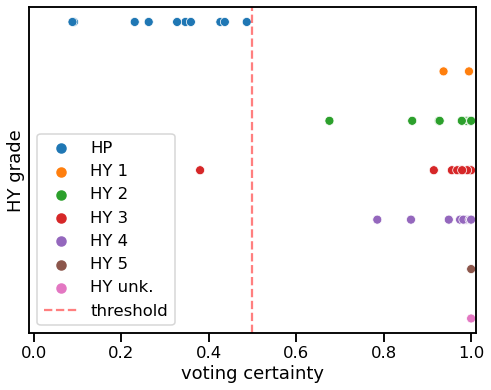}
	\caption{Plot of the voting certainty at different stages of Hoehn-Yahr scale.}
	\label{fig:voting}
\end{figure}

In Figure \ref{fig:voting} we present the voting inference metric. The measure is equivalent to the fraction of single-sample predictions that were predicted as PD-positive in a given patient. HP is healthy population. Dotted line is the 0.5 votes threshold between positive and negative grading. An important observation is that the only misclassified subject is a \emph{false negative}, which very undesired in a medical classification system. Additional metrics, including false negative rate, are shown in Table \ref{tab:results_tpr_tnr}.

\begin{center}
\begin{table}[h!]
\centering
\caption{\label{tab:results_tpr_tnr}Comparison of models' sensitivity and specificity}
\setlength{\arrayrulewidth}{1pt}
\begin{tabularx}{1\textwidth} { 
  | >{\centering\arraybackslash}X 
  | >{\centering\arraybackslash}X 
  | >{\centering\arraybackslash}X | }
 \hline
 Model & Inferred voting sensitivity (true positive rate) & Inferred voting specificity (true negative rate) \\ \hline \hline
 frozen conv  & 0.97 & \textbf{1.00} \\ \hline
 full + pretrained & \textbf{1.00} & 0.20 \\ \hline
 full + not pretrained & \textbf{1.00} & 0.20 \\ \hline
\end{tabularx}
\end{table}
\end{center}

In Table \ref{tab:results}, we compare single-sample accuracy to inferred voting accuracy across 3 training setups. Two observations can be drawn from the table:
\begin{enumerate}
    \item pretraining improves the classification performance
    \item fine-tuning the convolutional part degrades the classification performance
\end{enumerate}

\section{Human-level performance assessment and interpretability}

The aim the performed assessment was to determine if human experts can also \textit{pick up} some signal in speech recordings solely. A survey was conducted among experts in neurology who did not examine the patients otherwise. In a provided questionnaire, the experts were provided with the recordings sampled from different parts of the phonetic test.

The subsets in the questionnaire consisted of:
\begin{itemize}
    \item all parts of the phonetic test
    \item only full sentences
    \item only words and syllables
    \item only vowels and sustained phonation
\end{itemize}

Six experts took part in the survey. They were asked to label each set of recordings (per patient) with one of the following: {no symptoms; mild PD symptoms; advanced PD symptoms; symptoms of a disease other than PD}. We provide a detailed table of collected answers in Appendix \ref{app:experts}. Averaged accuracy of the experts predictions on a binary task of PD scores up to 75\% when using mode value (similar to the proposed voting inference). We can therefore draw a conclusion that: 1) our model outperforms the human experts in speech classification; 2) there is a significant signal that can be distilled from speech only. This encourages further examination of the model's explainability, which could provide experts with reliable diagnostic input and promote trust in the proposed AI-based diagnostic tool.

\begin{figure}[h!]
	\centering
	\includegraphics[width=0.8\textwidth]{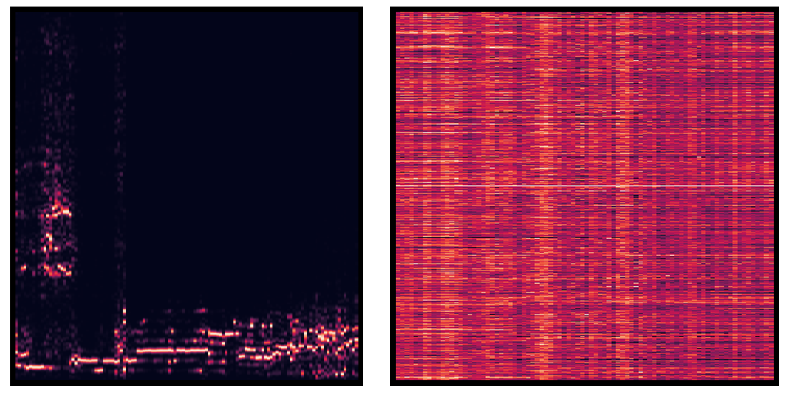}
	\caption{Comparison of a spectrogram (left) with a wav2vec feature map (right) for a sentence "dziś jest ładna pogoda" ("the weather is nice today"). Spectrogram visualises the acoustic input in time-frequency domain, while extracted feature map does so in a trainable time-features domain.}
	\label{fig:filters}
\end{figure}

We also approach the model in terms of interpretability. We wanted to observe if the feature map created by wav2vec convolutional layers can be interpreted in time-feature domain, similar to how spectrograms are analysed in time-frequency domain. We present a sample comparison in figure \ref{fig:filters}. The spectrogram was calculated with FFT length 1024 and 1/8 window overlap so that the output frequency resolution matched with the number of features in the internal wav2vec representation (512). We observe that there is no interpretable pattern in the feature map, however, the representation is much more evenly distributed than in case of the spectrogram, meaning that it likely yields more information.

\section{Discussion}

In our experiments, we have shown that it is possible to use an audio model trained on natural language to improve the performance on a downstream medical task. Our novel contribution is the construction of a machine learning framework for medical audio classification that takes the advantage of existing speech processing models. We have shown that our implementation obtains very good performance on downstream tasks, scoring up to 97.92\% accuracy.

In needs to be noted, however, that our approaches towards obtaining accurate multi-class predictions for different stages of the disease were so far unsuccessful, most probably due to insufficient representation of each Hoehn-Yahr subset in the training data. Further studies should focus on constructing a model that would differentiate the subjects in terms of the stage of disease's development. Probably, a different grading scale could be be used, such as UPDRS.

Our classifier should also be used with a given uncertainty margin, especially when considering an implementation of a downstream diagnostic tool. Our method can be used efficiently to separate healthy population from PD patients, but false negative rate has to be taken into account to avoid missing disease-impaired subjects. Another study should also check how the classifier performs in the presence of other diseases, most importantly ones impairing the human speech in any way. Having addressed all these uncertainties, it might be possible to develop a remote diagnostic tool for supporting the traditional clinical PD diagnostic process, based on the proposed method.

\section{Acknowledgements}
Model training was executed on ACK Cyfronet Prometheus cluster, using PLGrid infrastructure.

All of the patient’s data was collected and processed according to the decision of The Bioethics
Committee of the Jagiellonian University (decision no. 1072.6120.271.2019 from 21 November 2019)

\bibliographystyle{plainnat}
\bibliography{references}

\pagebreak


\appendix

\section{A detailed description of the phonetic test}
\label{app:phonetic_test}

The participants were asked to read out loud a phonetic test in Polish language \cite{maka_polish_2009} that consists of:
\begin{itemize}
  \item vowels \symbol{92}a, \symbol{92}e, \symbol{92}i, \symbol{92}u pronounced normally (3x)
  \item sustained phonation of vowels \symbol{92}a, \symbol{92}e, \symbol{92}i, \symbol{92}u (3x)
  \item words \{ala, as, ula, ela, igła\} (3x)
  \item sentences:
  \begin{itemize}
      \item Dziś jest ładna pogoda. (3x)
      \item Jacek mył kota.
      \item Lola lubi bal.
      \item Rysiek narysował bar.
      \item Marysia namalowała dym.
  \end{itemize}
\end{itemize}

Full recordings were later manually segmented into audio samples containing fragments of speech described above. The total length of the segmented speech samples was approximately 38 minutes in 2141 .wav files, giving on average 43 recordings per subject. Exact numbers vary between patients due to the manual quality check process which ruled out incomprehensible and noisy samples.

\section{Hoehn-Yahr stages in patients}
\label{app:HY}

Table \ref{tab:counts} presents how Hoehn-Yahr stages are distributed across the patients in the dataset.

\begin{center}
\begin{table}[h!]
\centering
\caption{\label{tab:counts}Value counts in target subgroups}
\setlength{\arrayrulewidth}{1pt}
\begin{tabularx}{1\textwidth} { 
  | >{\centering\arraybackslash}X 
  | >{\centering\arraybackslash}X 
  | >{\centering\arraybackslash}X
  | >{\centering\arraybackslash}X | }
 \hline
 \textbf{Group} & \textbf{Hoehn-Yahr grade} & \textbf{Count} & \textbf{Count (group)} \\ \hline \hline
 PD & 5 & 1  & 38 \\ & 4 & 11 & \\ & 3 & 13 &\\ & 2 & 11 & \\ & 1 & 2 & \\ \hline HP & \multicolumn{3}{c|}{10} \\ \hline
\end{tabularx}
\end{table}
\end{center}

\section{Audio augmentations}
\label{app:audiomentations}

\textbf{Addition of background noise.} Two noise recordings were used to be randomly sampled into the training set:
\begin{itemize}
    \item Recording of a busy street with people talking unintelligibly and objects rattling. Duration: 2:21 minutes. 
    \item Recording of street traffic with cars passing by at different speeds. Duration: 2:00 minutes.
\end{itemize}
Random fragments of background noise were sampled at every iteration and added to the training samples.

\textbf{Addition of coloured noise.} Parameters drawn randomly from:
\begin{itemize}
    \item signal-to-noise ratio (SNR) [dB] in range [3, 30]
    \item $f_{decay}$ in range [-2, 2]
\end{itemize}
\textbf{Shift.} Temporal shift was applied in range of $\pm$10\% difference without rollover. \newline
\textbf{Polarity inversion.} Applied to the whole training sample.

Each of the augmentations was applied with 50\% probability, drawn at every iteration for every augmentation separately.

\pagebreak

\section{Training logs}
\label{app:training}

Below, we present logged training losses for the simplified setup and the full transformer model.

\begin{figure}[h!]
	\centering
	\includegraphics[width=0.7\textwidth]{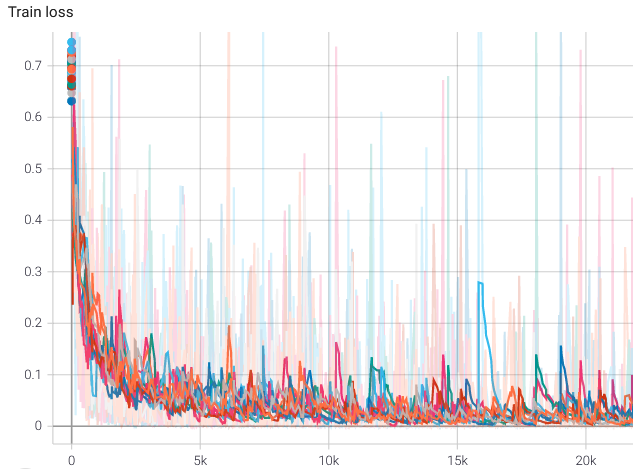}
	\caption{Training loss of several runs using the simplified model shown in \ref{fig:model}.}
	\label{fig:train_loss_frozen}
\end{figure}

\begin{figure}[h!]
	\centering
	\includegraphics[width=0.7\textwidth]{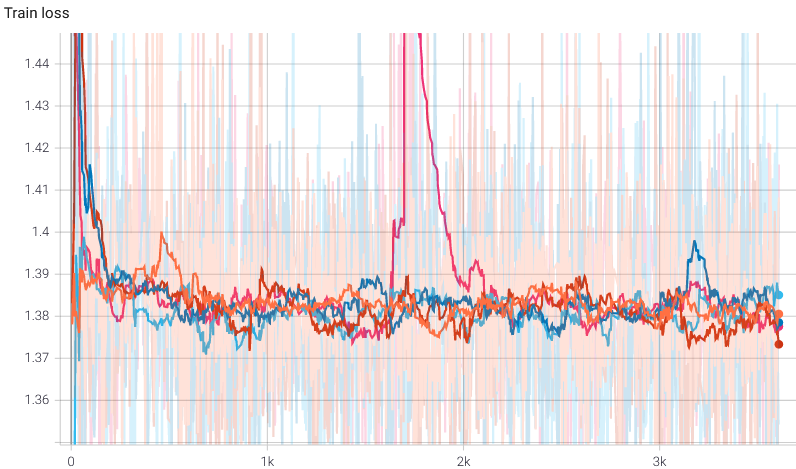}
	\caption{Training loss of several runs using the full wav2vec transformer. In all tested setups, the transformer model failed to converge.}
	\label{fig:train_loss_transformer}
\end{figure}

\pagebreak

\section{Expert questionnaire - detailed answers}
\label{app:experts}

\begin{center}
\begin{table}[ht]
\centering
\caption{\label{tab:questionnaire}Summary of the answers to the questionnaire. Options in the questionnaire were: 1 - no symptoms, 2 - symptoms other than PD, 3 - early-stage PD, 4 - advanced-stage PD. `Hit` means that at least one of the experts provided correct answer.}
\begin{tabularx}{0.9\textwidth} { 
  | >{\centering\arraybackslash}X 
  | >{\centering\arraybackslash}X 
  | >{\centering\arraybackslash}X
  | >{\centering\arraybackslash}X
  | >{\centering\arraybackslash}X |}
 \hline
 \multicolumn{5}{|c|}{\textbf{Set 1: all types of recordings}} \\
 \hline
 Subject & True H-Y & Mode(s) & Average & Hit \\
 \hline
 1 (PD) & 1 & 1 & 1.3 & YES  \\ \hline
 2 (HP) & - & 1, 3 & 2.0 & YES \\ \hline
 3 (PD) & 3 & 3 & 3.0 & YES \\ \hline
 4 (HP) & - & 1 & 1.0 & YES \\ \hline
 5 (PD) & 5 & 2, 3 & 2.5 & YES \\ \hline
 6 (HP) & - & 1 & 1.7 & YES \\ \hline
 \hline
  \multicolumn{5}{|c|}{\textbf{Set 2: only full sentences}} \\
 \hline
 Subject & True H-Y & Mode(s) & Average & Hit \\
 \hline
 7 (PD) & 2 & 1 & 1.3 & YES  \\ \hline
 8 (PD) & 2 & 3 & 2.3 & YES \\ \hline
 9 (PD) & 4 & 4 & 3.3 & YES \\ \hline
 10 (PD) & 4 & 4 & 3.3 & YES \\ \hline
 11 (HP) & - & 1 & 1.0 & YES \\ \hline
 12 (HP) & - & 3 & 2.7 & YES \\ \hline
 \hline
  \multicolumn{5}{|c|}{\textbf{Set 3: words and syllables}} \\
 \hline
 Subject & True H-Y & Mode(s) & Average & Hit \\
 \hline
 13 (HP) & - & 3 & 3.2 & NO  \\ \hline
 14 (PD) & 2 & 1 & 2.3 & YES \\ \hline
 15 (HP) & - & 3 & 2.7 & YES \\ \hline
 16 (HP) & - & 1 & 1.5 & YES \\ \hline
 17 (HP) & - & 1 & 1.3 & YES \\ \hline
 18 (PD) & 4 & 4 & 3.2 & YES \\ \hline
 \hline
 \multicolumn{5}{|c|}{\textbf{Set 4: vowels and sustained phonation}} \\
 \hline
 Subject & True H-Y & Mode(s) & Average & Hit \\
 \hline
 19 (PD) & 2 & 4 & 3.2 & YES \\ \hline
 20 (HP) & - & 1 & 1.5 & YES \\ \hline
 21 (PD) & 3 & 4 & 3.7 & NO \\ \hline
 22 (PD) & 4 & 2, 4 & 3.0 & YES \\ \hline
 23 (PD) & 4 & 4 & 4.0 & YES \\ \hline
 24 (PD) & 2 & 4 & 4.0 & YES \\ \hline
\end{tabularx}
\end{table}
\end{center}

\end{document}